\documentclass[letterpaper, 10 pt, conference]{ieeeconf}  

\IEEEoverridecommandlockouts                              

\usepackage{amsmath,amssymb,graphicx}
\usepackage{float, booktabs}
\usepackage{multirow}
\usepackage{pifont}
\newcommand{\cmark}{\ding{51}} 
\newcommand{\xmark}{\ding{55}} 

\usepackage[
backend=biber,
sorting=none
]{biblatex}
\title{\LARGE \bf
Dynamic Contextual Attention Network: Transforming Spatial Representations into Adaptive Insights for Endoscopic Polyp Diagnosis
}
\author{Teja Krishna Cherukuri$^{1*}$, Nagur Shareef Shaik$^{1*}$, Sribhuvan Reddy Yellu$^{1}$, Jun-Won Chung$^{2}$, Dong Hye Ye$^{1\dag}$ \\ $^1$Department of Computer Science, Georgia State University, $^2$Gachon University, Gil Medical Center
\thanks{$^{*}$Equal Contribution}
\thanks{$^{\dag}$Corresponding Author: \tt\small dongye@gsu.edu}
}

\begin{document}

\maketitle
\thispagestyle{empty}
\pagestyle{empty}

\begin{abstract}
Colorectal polyps are key indicators for early detection of colorectal cancer. However, traditional endoscopic imaging often struggles with accurate polyp localization and lacks comprehensive contextual awareness, which can limit the explainability of diagnoses. To address these issues, we propose the Dynamic Contextual Attention Network (DCAN). This novel approach transforms spatial representations into adaptive contextual insights, using an attention mechanism that enhances focus on critical polyp regions without explicit localization modules. By integrating contextual awareness into the classification process, DCAN improves decision interpretability and overall diagnostic performance. This advancement in imaging could lead to more reliable colorectal cancer detection, enabling better patient outcomes.
\end{abstract}

\begin{keywords}
Dynamic Contextual Attention, Colorectal Polyps, Endoscopic Imaging, Explainable Diagnosis
\end{keywords}

\section{Introduction}
\label{sec:intro}

Colorectal cancer poses a significant global health challenge, with colorectal polyps serving as critical precursors that necessitate timely identification and intervention \cite{li2018recent}. Despite advancements in medical imaging and diagnostics, current endoscopic techniques often struggle to accurately localize and classify polyps while providing sufficient explainability \cite{fu2021future}. The inherent variability in polyp appearance, including differences in size, shape, texture, and illumination noise, complicates detection efforts, often resulting in missed diagnoses and delays in treatment \cite{gupta2024systematic}. This drives the need for enhanced diagnostic tools that assist clinicians in making informed decisions while minimizing the risk of human error during colonoscopy procedures. Recent studies have explored various deep learning methods to improve polyp detection and classification \cite{chou2023improving}. For instance, the Tiny Polyp detection method leverages Vision Transformers to enhance feature extraction while maintaining computational efficiency, significantly improving recall and precision compared to traditional models like YOLOv5 \cite{liu2024tiny}. Similarly, the YOLO-V8 network has demonstrated remarkable results in achieving high precision and recall rates in polyp detection through a robust AI-driven approach, indicating the potential for integrating advanced neural networks into clinical workflows \cite{lalinia2024colorectal}. Other methods, such as deep convolutional neural networks (CNNs), have focused on addressing challenges posed by polyp variability, employing innovative techniques for feature extraction and data augmentation to enhance model performance \cite{rahim2021deep}. However, despite these advancements, existing methods exhibit notable drawbacks. Many traditional models struggle with explainability and the integration of contextual information, leading to difficulties in clinical acceptance \cite{ahamed2024automated}.

Attention mechanisms, including Spatial, Squeeze \& Excitation, Non-local, Global Context, Gated, Self-Attention, Triplet Attention, and Guided Context Gating, are designed to enhance model performance by prioritizing relevant features. However, they frequently exhibit limitations in polyp diagnosis from endoscopic images, which may hinder their effectiveness in clinical applications. For example, Spatial Attention may overlook critical local relationships between anatomical structures, leading to misinterpretations \cite{shaik2022multi}, while Squeeze \& Excitation improves feature representation but fails to capture nuanced interactions essential for accuracy \cite{hu2018squeeze}. Non-local \cite{wang2018non} and Global Context \cite{cao2020global} attention mechanisms can introduce noise, complicating the extraction of clinically relevant insights. Gated \cite{jo2019attention} and Self-Attention \cite{NIPS2017_3f5ee243} mechanisms dynamically adjust feature importance but lack robustness in managing polyp variability, which is crucial in clinical settings. Triplet Attention \cite{misra2021rotate}, which leverages cross-dimensional interactions to capture spatial and channel-wise dependencies, offers a more comprehensive feature representation compared to Self-Attention. However, it may still struggle with the inherent variability and noise in endoscopic images, particularly under challenging illumination conditions. Furthermore, Guided Context Gating \cite{cherukuri2024guided}, while enhancing localized lesion context, may be biased towards induced light illuminations, distorting attention to significant anatomical features. These limitations underscore the necessity for an attention framework that not only enhances focus on relevant features but also refines illumination noise, providing clear and interpretable insights to support reliable medical decision-making.

In response to these challenges, we propose the Dynamic Contextual Attention Network (DCAN), which aims to bridge the gap between detection performance and explainability in polyp classification. Our proposed attention mechanism effectively transforms spatial representations into adaptive contextual insights, allowing the model to enhance its focus on potential polyp regions without relying on explicit localization modules. This approach enables the dynamic adaptation of diagnostically relevant regions, thereby improving the explainability of the model's decisions.

\begin{figure*}
    \centering
    \includegraphics[width=.9\linewidth]{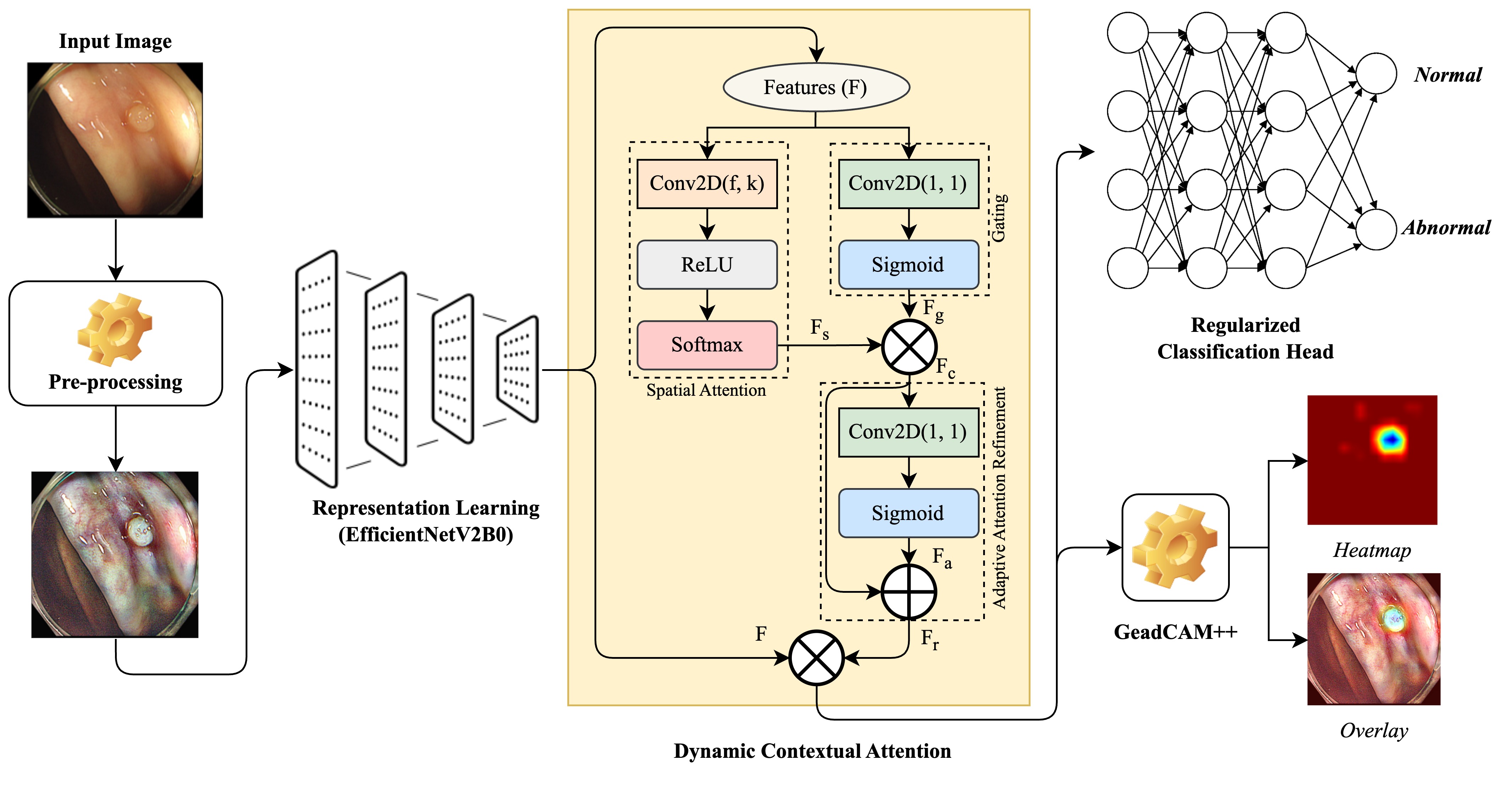}
    \vspace{-10pt}
    \caption{Architecture of the Proposed Dynamic Contextual Attention Network (DCAN). Pre-processing applies CLAHE to enhance contrast. EfficientNet$V_2B_0$ extracts spatial features, while the Dynamic Contextual Attention mechanism boosts diagnosis accuracy and explainability by focusing on relevant anatomical regions and filtering out noise. The Regularized Classification Head categorizes images as Normal or Abnormal, and GradCAM++ visually explains the DCAN's decisions.}
    \vspace{-10pt}
    \label{fig:DACN_Architecture}
\end{figure*}

\section{Methodology} \label{sec:methods}

The primary objective of this research is to design a colorectal polyp detection framework that significantly enhances both detection accuracy and explainability by dynamically targeting relevant anatomical regions while effectively filtering out noise from inconsistent illumination and irrelevant structures - common challenges in endoscopic images. To achieve this, we develop and evaluate the Dynamic Contextual Attention Network (DCAN), an advanced attention-based approach for colorectal polyp detection. Our focus is on improving polyp localization without relying on explicit localization modules through the multi-stage attention process of DCA, which employs spatial, gated, and adaptive attention refinement mechanisms to dynamically concentrate on diagnostically relevant regions. This approach aims to reduce miss detections caused by variations in polyp size, shape, and texture. The following subsections details the components of the proposed method.

\subsection{Representation Learning} 

Representation learning is crucial for effectively capturing complex patterns from medical images. For polyp detection, leveraging pre-trained models allows us to build upon existing knowledge and enhance the performance of our framework. In this study, we employ the pre-trained EfficientNet$V_2B_0$ model to extract initial feature representations from the pre-processed endoscopic images, leveraging its efficient architecture and robust performance to minimize computational overhead \cite{tan2021efficientnetv2}. Given an image \(I\), fine-tuning the EfficientNet$V_2B_0$ model on our curated dataset of endoscopic images yields feature maps of dimensions \((16 \times 16 \times 1280)\), represented as \(F = \text{EfficientNet}{V_2B_0}(I)\). This results in a compact representation that captures 1280 distinct features distributed across a \(16 \times 16\) grid, effectively encapsulating vital information necessary for precise polyp detection and localization in subsequent processing stages. 

\subsection{Dynamic Contextual Attention}

The goal of the Dynamic Contextual Attention (DCA) is to enhance feature representation in convolutional neural networks (CNNs) by dynamically focusing on relevant spatial regions while filtering out irrelevant or noisy features. Traditional attention mechanisms often highlight coarse regions without considering finer distinctions or contextual dependencies between spatial locations. The DCA mechanism overcomes these limitations by combining spatial, gating, and adaptive attention refinement mechanisms to capture more refined, contextually aware attention maps. This approach is designed to improve tasks that require fine-grained feature localization in endoscopic imaging, where precise attention to detail is critical.

\subsubsection{Spatial Attention} 

The spatial attention focuses on relevant spatial locations within the input feature map and highlights regions of interest. Given an input feature map \(F \in \mathbb{R}^{H \times W \times D}\), where \(H\), \(W\), and \(D\) denote the height, width, and number of features respectively, we first apply a convolutional layer with kernel size \(k \times k\), number of filters \(F\), and stride \(s\) to extract spatial features. The resulting spatial featurare \(F_{\text{spatial}}\) is then activated through a ReLU and normalized using a softmax operation over the spatial dimensions:

\begin{equation}
F_{\text{s}} = \text{Softmax}(\left(\Gamma(W_s * F + b_s)\right)
\label{eq:spatial_attention}
\end{equation}

In Eq. \ref{eq:spatial_attention}, \( W_s \) and \( b_s \) represent the weights and biases of the convolutional layer, \( * \) denotes the convolution operation, and \( \Gamma \) is the ReLU activation function. The softmax function ensures that regions most likely to contain polyps receive higher attention scores, helping the network isolate regions of interest from irrelevant background information.

\subsubsection{Gating}

Gating mechanism adds an additional layer of filtering that learns to amplify important spatial regions while downweighting irrelevant ones. The gating mechanism is designed to capture global dependencies, controlling the flow of features and allowing the network to selectively ignore noise. This is achieved via a 1x1 convolution followed by a sigmoid activation as represented in Eq. \ref{eq:gating_attention}.

\begin{equation}
F_{\text{g}} = \text{Sigmoid}(W_g * F + b_g)
\label{eq:gating_attention}
\end{equation}

Here, \( W_g \) and \( b_g \) are the weights and biases of the gating convolution. The sigmoid activation suppresses features unrelated to polyp regions, allowing the model to selectively attend to polyp regions while filtering out background features. The spatial and gating attention maps are combined by performing element-wise multiplication, $F_{\text{c}} = F_{\text{s}} \odot F_{\text{g}}$. The combined attention map \( F_{\text{c}} \) leverages the strengths of both spatial attention and gating mechanisms and allows the model to focus on polyp regions based on spatial context while also filtering out irrelevant features. 

\subsubsection{Adaptive Attention Refinement}

An additional adaptive refinement step is introduced to further enhance focus on relevant regions while suppressing residual noise. This is achieved through a convolution operation followed by a sigmoid activation, as shown in Eq. \ref{eq:adaptive_attention}. The final attention map is computed by adding the adaptive attention weights to the combined spatial and gating attention maps (Eq. \ref{eq:refined_attention}). Lastly, the refined attention map is multiplied element-wise with the original feature maps to generate the attended feature maps (Eq. \ref{eq:attended_features}). This multi-stage refinement ensures that the network robustly identifies polyp regions while minimizing lighting artifacts and irrelevant noise, leading to improved localization and diagnostic insights.

\begin{equation}
F_{\text{a}} = \sigma(W_a * F + b_a \label{eq:adaptive_attention}) 
\end{equation}
\begin{equation}
F_{\text{r}} = F_{\text{c}} + F_{\text{a}} \label{eq:refined_attention}
\end{equation}
\begin{equation}
F_{\text{dca}} = F_{\text{r}} \odot F \label{eq:attended_features}
\end{equation}

Where \( F_{\text{a}} \) is the adaptive refinement attention map generated through a convolutional layer with weights \( W_a \) and bias \( b_a \), followed by a sigmoid activation \( \sigma \), which ensures values are within the range [0, 1].  \( F \) represents the input feature maps from the backbone network, capturing the spatial representation of the image. \( F_{\text{r}} \) is the final refined attention map, obtained by combining the previously calculated attention map \( F_{\text{c}} \) with the adaptive attention map \( F_{\text{a}} \).  \( F_{\text{dca}} \) denotes the final attended feature maps, resulting from the element-wise multiplication \( \odot \) between the refined attention map \( F_{\text{r}} \) and the original feature maps \( F \). This multiplication applies the learned attention to the spatial features, emphasizing polyp regions while suppressing irrelevant illuminations.

\subsection{Regularized Classification Head}

To achieve robust classification performance in our framework, we implement a regularized classification head following the Dynamic Contextual Attention (DCA) mechanism. This includes a Global Average Pooling (GAP) operation that aggregates the attended feature maps \(F_{\text{dca}} \in \mathbb{R}^{H \times W \times D}\) into a one-dimensional vector, effectively capturing the average features across spatial dimensions. Subsequently, a dense layer processes this pooled output to extract complex representations, while dropout regularization is applied introduing stochasticity, aiding in regularization by randomly deactivating half of the neurons during training to mitigate the risk of overfitting. Finally, the output layer generates class prediction probabilities using the softmax activation function.

\begin{equation}
F_\text{gap} = \frac{1}{H \times W} \sum_{i=1}^{H} \sum_{j=1}^{W} f_{ij} \quad \forall f_{ij} in F_{\text{dca}} \label{eq:gap}.
\end{equation}
\begin{equation}
F_{\text{d}} = \Gamma(W_{\text{1}} \cdot F_\text{gap} + b_{\text{1}}) \label{eq:dense}.
\end{equation}
\begin{equation}
F_{\text{do}} = \text{Dropout}(0.3)(F_{\text{d}}) \label{eq:dropout}.
\end{equation}
\begin{equation}
\text{C} = \text{Softmax}(W_{\text{2}} \cdot F_{\text{do}} + b_{\text{2}}) \label{eq:output}.
\end{equation}

Equations \ref{eq:gap} to \ref{eq:output} illustrates the operations involved in this module where, \(W_{\text{1}}, W_{\text{2}}\) and \(b_{\text{1}}, b_{\text{2}}\) represent the weights and biases of the dense layers, while \(\Gamma\) denotes the ReLU activation. All module parameters undergo L2 normalization, constraining weight vectors to a unit norm. 

The model is trained end-to-end, encompassing all modules discussed, utilizing cross-entropy loss defined as $\mathcal{L} = -\sum_{i=1}^{N} \sum_{c=1}^{C} y_{i,c} \log(p_{i,c})$, where \(N\) is the number of samples, \(C\) is the number of classes, \(y_{i,c}\) is the ground truth label for sample \(i\) and class \(c\), and \(p_{i,c}\) is the predicted probability of sample \(i\) belonging to class \(c\). The parameters are updated using the AdamW optimizer, which incorporates weight decay in its update rule $\theta_{t+1} = \theta_t - \eta \left( m_t \odot \frac{1}{\sqrt{v_t} + \epsilon} + \lambda \theta_t \right)$, where \(\theta_t\) are the model parameters (weights and biases) at \(t\), \(\eta\) is the learning rate, \(m_t\) and \(v_t\) are the first and second moments of the gradients, \(\epsilon\) is a small constant to prevent division by zero, and \(\lambda\) is the weight decay coefficient. This comprehensive training approach facilitates effective learning and generalization in the context of polyp detection.

\section{Experimental Results} \label{sec:results}

\subsection{Datasets} 

We employed two datasets to evaluate and benchmark the proposed approach: an in-house dataset and a publicly available dataset. The in-house dataset, curated from Gachon University Gil Medical Center, comprises 1,070 retrospectively collected labeled endoscopic images, including 550 abnormal cases with polyps and 520 normal cases without polyps. This dataset was specifically designed to assess the performance of polyp detection and diagnosis models in a clinical setting. For broader validation, we utilized the HyperKvasir dataset \cite{borgli2020hyperkvasir}, a large-scale publicly available gastrointestinal (GI) tract image and video dataset. Collected using Olympus and Pentax endoscopy equipment at Bærum Hospital, Norway, HyperKvasir includes 10,662 labeled images spanning 23 distinct GI tract findings, such as polyps, esophagitis, and ulcers. This dataset is widely recognized as a benchmark for evaluating GI tract endoscopic image analysis methods and provides a diverse range of cases to test the generalizability of the proposed approach. 

\begin{table*}[!ht]
    \centering
    \caption{Comparing the performance of various Attention mechanisms against proposed attention on GMC Polyp Dataset}
    \label{tab:attention_metrics}
    \setlength{\tabcolsep}{3pt}
    \begin{tabular}{lccccc}
        \toprule
        \textbf{Attention} & \textbf{Accuracy (\%)} & \textbf{Precision (\%)} & \textbf{Recall (\%)} & \textbf{F1-Score (\%)} & \textbf{Kappa (\%)} \\
        \midrule
        No Attention \cite{tan2021efficientnetv2} & $83.80 \pm 1.50$ & $84.20 \pm 1.40$ & $83.80 \pm 1.50$ & $83.80 \pm 1.60$ & $67.30 \pm 3.30$ \\
        Spatial Attention \cite{shaik2022multi} & $91.68 \pm 2.13$ & $91.73 \pm 2.16$ & $91.72 \pm 2.17$ & $91.66 \pm 2.13$ & $83.33 \pm 4.27$ \\
        Squeeze \& Excitation \cite{hu2018squeeze} & $91.21 \pm 3.73$ & $91.34 \pm 3.64$ & $91.33 \pm 3.68$ & $91.20 \pm 3.73$ & $82.42 \pm 7.44$ \\
        Gated Attention \cite{jo2019attention} & $88.94 \pm 0.54$ & $94.89 \pm 0.90$ & $84.56 \pm 0.66$ & $89.43 \pm 0.50$ & $77.92 \pm 1.09$ \\
        Global Context Attention \cite{cao2020global} & $89.43 \pm 3.91$ & $89.51 \pm 3.83$ & $89.51 \pm 3.98$ & $89.41 \pm 3.93$ & $78.85 \pm 7.85$ \\
        Non-Local Attention \cite{wang2018non} & $85.14 \pm 3.22$ & $85.62 \pm 2.95$ & $85.11 \pm 3.05$ & $85.04 \pm 3.22$ & $70.23 \pm 6.34$ \\
        Self-Attention \cite{NIPS2017_3f5ee243} & $89.25 \pm 5.17$ & $89.41 \pm 5.09$ & $89.35 \pm 5.18$ & $89.23 \pm 5.16$ & $78.50 \pm 10.31$ \\
        Triplet Attention \cite{misra2021rotate} & $93.47 \pm 1.21$ & $93.41 \pm 1.21$ & $93.41 \pm 1.20$ & $93.37 \pm 1.21$ & $95.25 \pm 0.80$ \\ 
        Guided Context Gating \cite{cherukuri2024guided} & $93.26 \pm 0.62$ & $93.57 \pm 0.91$ & $92.98 \pm 1.12$ & $93.27 \pm 1.01$ & $87.75.59 \pm 2.42$ \\
    \textbf{Dynamic Context Attention (Ours)} & \textbf{94.75 $\pm$ 0.79} & \textbf{94.81 $\pm$ 0.81} & \textbf{94.79 $\pm$ 0.81} & \textbf{94.75 $\pm$ 0.79} & \textbf{89.50 $\pm$ 1.59} \\
        \bottomrule
    \end{tabular}
    \vspace{-3mm}
\end{table*}

\begin{table*}[!ht]
    \centering
    \caption{Comparing the performance of various Attention mechanisms against proposed attention on HyperKvasir Dataset}
    \label{tab:attention_metrics_hyper-kvasir}
    \setlength{\tabcolsep}{3pt}
    \begin{tabular}{lccccc}
        \toprule
        \textbf{Attention} & \textbf{Accuracy (\%)} & \textbf{Precision (\%)} & \textbf{Recall (\%)} & \textbf{F1-Score (\%)} & \textbf{Kappa (\%)} \\
        \midrule
        No Attention \cite{tan2021efficientnetv2} & $84.63 \pm 2.49$ & $83.58 \pm 1.82$ & $84.63 \pm 2.49$ & $83.94 \pm 2.30$ & $93.20 \pm 0.84$ \\
        Spatial Attention \cite{shaik2022multi} & $89.06 \pm 0.42$ & $88.48 \pm 0.51$ & $89.06 \pm 0.42$ & $88.63 \pm 0.47$ & $95.25 \pm 0.14$ \\
        Squeeze \& Excitation \cite{hu2018squeeze} & $89.09 \pm 0.45$ & $88.80 \pm 0.21$ & $89.09 \pm 0.45$ & $88.81 \pm 0.47$ & $95.37 \pm 0.68$ \\
        Gated Attention \cite{jo2019attention} & $88.70 \pm 0.22$ & $88.73 \pm 0.29$ & $88.70 \pm 0.22$ & $88.62 \pm 0.25$ & $94.88 \pm 0.25$ \\
        Global Context Attention \cite{cao2020global} & $89.25 \pm 0.06$ & $88.85 \pm 0.11$ & $88.75 \pm 0.65$ & $88.88 \pm 0.15$ & $95.41 \pm 0.28$ \\
        Non-Local Attention \cite{wang2018non} & $87.63 \pm 2.49$ & $86.58 \pm 1.82$ & $87.63 \pm 2.49$ & $86.94 \pm 2.30$ & $94.20 \pm 0.84$ \\
        Self-Attention \cite{NIPS2017_3f5ee243} & $88.77 \pm 0.13$ & $88.65 \pm 0.24$ & $88.77 \pm 0.13$ & $88.56 \pm 0.13$ & $95.25 \pm 0.80$ \\ 
        Triplet Attention \cite{misra2021rotate} & $89.09 \pm 0.33$ & $88.97 \pm 0.24$ & $89.09 \pm 0.33$ & $89.07 \pm 0.33$ & $95.11 \pm 0.21$ \\
        Guided Context Gating \cite{cherukuri2024guided} & $89.27 \pm 0.31$ & $88.63 \pm 0.41$ & $89.27 \pm 0.31$ & $88.74 \pm 0.37$ & $95.47 \pm 0.09$ \\
        \textbf{Dynamic Context Attention (Ours)}  & $\textbf{89.54} \pm \textbf{0.48}$ & $\textbf{89.11} \pm \textbf{0.40}$ & $\textbf{89.54} \pm \textbf{0.48}$ & $\textbf{89.22} \pm \textbf{0.40}$ & $\textbf{95.34} \pm \textbf{1.03}$ \\
        \bottomrule
    \end{tabular}
    \vspace{-3mm}
\end{table*}

\begin{table*}[!ht]
    \centering
    \caption{Ablation Study: Performance comparison of including and excluding different components of proposed Attention}
    \label{tab:ablation_study}
    \begin{tabular}{ccccccccc}
        \toprule
        \textbf{Spatial} & \textbf{Gated} & \textbf{Refinement} & \textbf{Accuracy (\%)} & \textbf{Precision (\%)} & \textbf{Recall (\%)} & \textbf{F1-Score (\%)} & \textbf{Kappa (\%)} \\
        \midrule
        \cmark & \xmark & \xmark & $92.99 \pm 1.26$ & $93.12 \pm 1.26$ & $93.05 \pm 1.26$ & $92.99 \pm 1.26$ & $87.00 \pm 2.53$ \\
        \xmark & \cmark & \xmark & $93.09 \pm 1.63$ & $93.26 \pm 1.50$ & $93.11 \pm 1.66$ & $93.08 \pm 1.64$ & $87.18 \pm 3.27$ \\
        \cmark & \cmark & \cmark & $\mathbf{94.75 \pm 0.79}$ & $\mathbf{94.81 \pm 0.81}$ & $\mathbf{94.79 \pm 0.81}$ & $\mathbf{94.75 \pm 0.79}$ & $\mathbf{89.50 \pm 1.59}$ \\
        \bottomrule
    \end{tabular}
\end{table*}

\subsection{Implementation Details}

All endoscopic images were resized to dimensions of $(512 \times 512 \times 3)$ and applied Contrast Limited Adaptive Histogram Equalization (CLAHE) \cite{pizer1990contrast} to improves local contrast and reduces the impact of illumination variations, making polyps more discernible in the images. Evaluating the robustness of propopsed approach, the in-house dataset was validated using a 5-fold cross-validation strategy and HyperKvasir was validated using standard 2-fold cross-validation protocol, aligning with prior research and enabling fair comparisons with existing methods.

\begin{figure*}[htbp]
    \centering
    \includegraphics[width=\textwidth]{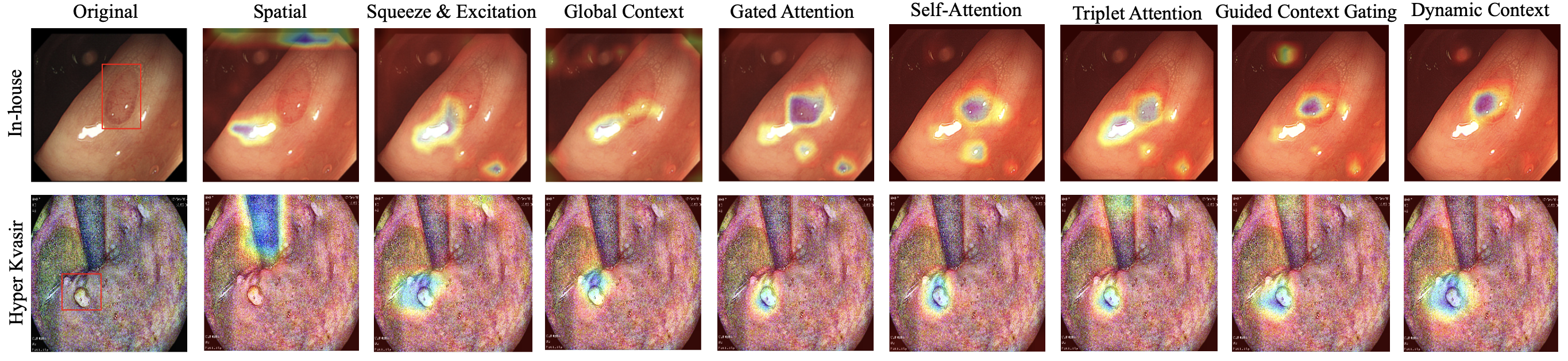}
    \vspace{-20pt}
    \caption{Attention maps generated by different attention mechanisms highlight crucial regions for polyp detection in endoscopic images from both the In-house (top) and HyperKvasir (bottom) datasets. The figure presents both non-CLAHE and CLAHE-enhanced images, showcasing their impact on the Dynamic Contextual Attention (DCA) model. A red-colored bounding box precisely localizes the polyp within the original endoscopic image.}
    \vspace{-5pt}
    \label{fig:attention_maps}
\end{figure*}

\begin{figure}[!ht]
    \centering
    \includegraphics[width=0.5\textwidth]{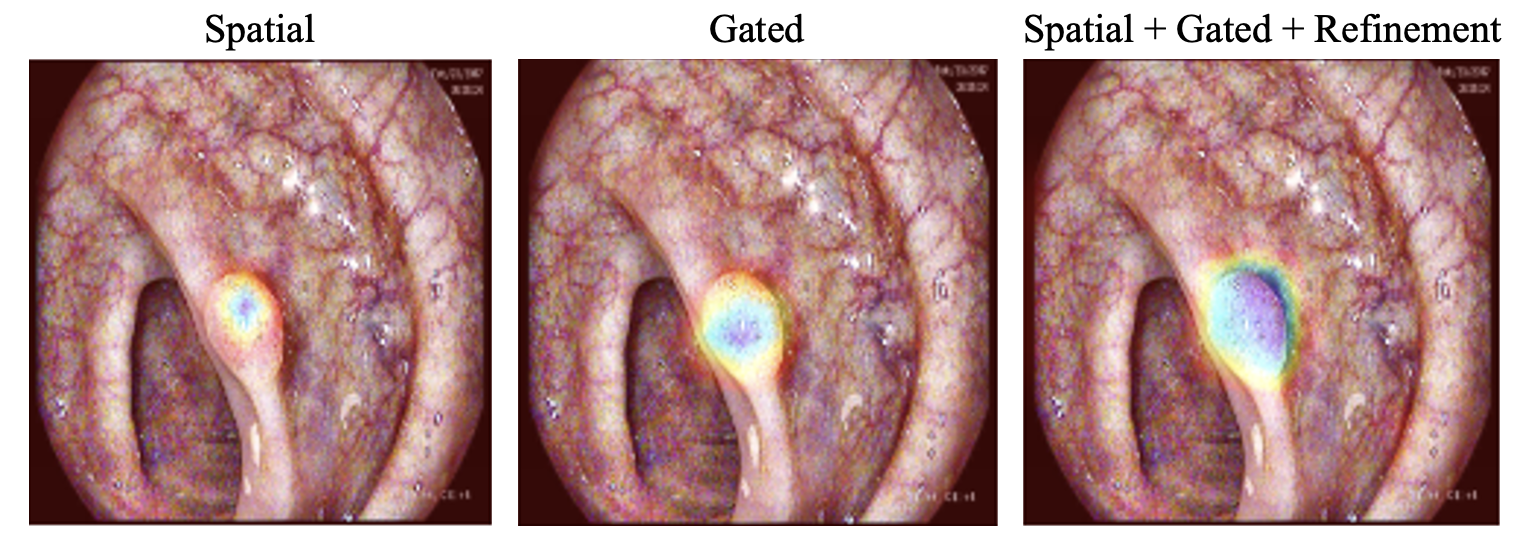}
    \vspace{-20pt}
    \caption{ Attention maps for polyp detection in abnormal class images post-CLAHE; The maps highlight the contribution of the Spatial and Gated components in conjunction with the Refinement component, demonstrating the importance of covering the entire polyp.}
    \vspace{-5pt}
    \label{fig:ablation-attention_maps}
\end{figure}

\subsection{Quantitative Evaluation}

The performance of various attention mechanisms was evaluated on both the in-house and the HyperKvasir datasets, as summarized in Tables \ref{tab:attention_metrics} and \ref{tab:attention_metrics_hyper-kvasir}, respectively. The proposed Dynamic Context Attention (DCA) consistently outperformed existing attention mechanisms across all metrics, demonstrating its effectiveness in polyp detection tasks. Initially, we conducted experiments with various pre-trained convolutional bases to determine the most effective backbone for polyp detection, out of which EfficientNet $V_2B_0$ emerged as the most promising feature extractor.

The comparative analysis of various attention mechanisms for polyp detection presented in table \ref{tab:attention_metrics} highlights the superior performance of the proposed Attention (DCA) method. DCA consistently outperforms traditional and contemporary attention models across all evaluated metrics, demonstrating notable improvements in-terms of all metrics. This analysis of reveals the pronounced superiority of the proposed Dynamic Context Attention (DCA) achieving up to a 1.49 \% improvement in accuracy over the strongest baseline Guided Context Gating, which reflects its superior capability in capturing essential contextual information. Furthermore, when compared to the traditional No Attention baseline, DCA showcases an impressive 10.95\% increase in accuracy, highlighting its efficacy in enhancing detection performance. This significant advancement emphasizes the potential of DCA to set a new standard in attention-based models for medical imaging, particularly in polyp detection tasks.

The results of experiments on HyperKvasir dataset presented in the table \ref{tab:attention_metrics_hyper-kvasir} proves the robustness of DCA achieving an accuracy of 89.54\%, outperforming Guided Context Gating and other baselines. While Triplet Attention and Guided Context Gating showed competitive results, DCA's consistent performance across both datasets highlights its ability to handle variability in polyp appearance and imaging conditions. Specifically, DCA demonstrated exceptional reliability and consistency in polyp class detection in HyperKvasir dataset, achieving a near-perfect precision of 99.90\% ($\pm$0.14\%), effectively minimizing false positives while maintaining high sensitivity. The recall of 97.28\% ($\pm$0.28\%) highlights its ability to accurately identify polyps across diverse imaging conditions, ensuring minimal missed detections. Notably, the low variance across multiple folds underscores its robustness, proving its stability even in the presence of variations in polyp appearance. This consistency not only reinforces DCA’s superiority in handling medical image complexities but also establishes it as a highly dependable approach for polyp detection, setting a new benchmark in medical image analysis.

\subsubsection{Ablation Study}

To validate the contribution of each component in the proposed Dynamic Context Attention (DCA), an ablation study was conducted on the in-house dataset, and the results are reported in Table \ref{tab:ablation_study}. The study evaluated the impact of the Spatial Attention, Gated Attention, and Refinement modules individually and in combination. When only Spatial Attention was used, the model achieved an accuracy of 92.99\%, as it effectively captures local spatial relationships between anatomical structures, which are crucial for polyp localization. Gated Attention alone yielded 93.09\%, demonstrating its ability to dynamically adjust feature importance and filter out irrelevant information, thereby enhancing feature representation. However, the integration of all three components—Spatial Attention, Gated Attention, and Refinement—resulted in the highest performance, achieving an accuracy of 94.75\%.

The Refinement module plays a pivotal role in this improvement by iteratively refining the attention maps, ensuring that the model focuses on the most relevant regions while suppressing noise and illumination artifacts. This module complements Spatial Attention by enhancing its ability to capture fine-grained details and Gated Attention by providing a more robust mechanism for feature selection. The combined effect of these components allows the model to better handle the variability in polyp appearance, shape, and size, as well as the challenges posed by uneven illumination and complex anatomical backgrounds. This synergy leads to a significant improvement in performance in terms of all reported metrics, and robustness, making the proposed DCA a highly effective solution for polyp detection in endoscopic images.

\subsection{Qualitative Results}

The GradCAM++ visualizations presented in the figure \ref{fig:attention_maps} show how different attention mechanisms highlight regions for polyp detection in endoscopic images. The Dynamic Contextual Attention (DCA) mechanism stands out by focusing directly on the polyp region for both In-house and HyperKvasir datasets, with minimal interference from surrounding tissues or reflections caused by lighting, making it highly robust to variations in illumination and external clip. In contrast, other attention methods, such as Spatial and Self-Attention, are distracted by bright areas or reflections and external clip, leading to noise and deviation in the attention maps. Similarly, Squeeze \& Excitation, Global Context and Triplet Attention mechanisms distribute attention across multiple irrelevant regions, diluting the focus on the polyp itself. Gated Attention and Guided Context Gating improve over the former by narrowing the attention field, yet they still exhibit some vulnerability to lighting artifacts. The DCA’s precision in highlighting only the relevant anatomical areas indicates its superior ability to filter out noise and detect polyps under various conditions, thus improving both detection accuracy and explainability.

Additionally, attention maps from Figure \ref{fig:ablation-attention_maps} illustrate the contribution of each component to the interpretability of endoscopic images. While Spatial Attention effectively localized the polyp region, and Gated Attention captured structural details by dynamically adjusting feature importance, each component alone had limitations. Spatial Attention struggled with fine-grained refinement, and Gated Attention occasionally included irrelevant regions due to its focus on global context. However, the integration of these components with the Adaptive Attention Refinement module addressed their individual weaknesses. The Adaptive Attention Refinement module iteratively refined the attention maps, ensuring precise and complete highlighting of the polyp region while suppressing noise and irrelevant artifacts. This synergy within the DCA block combines the strengths of Spatial Attention's localization capability, Gated Attention's dynamic feature selection, and Refinement's iterative precision, resulting in a robust and interpretable framework. Together, these components complement each other, overcoming individual limitations and delivering superior performance in polyp detection and visualization.

\section{Conclusion} \label{sec:conclusion}

In this work, we presented the Dynamic Contextual Attention Network (DCAN), a novel framework designed to enhance polyp detection and classification in endoscopic imaging. By incorporating dynamic attention mechanisms, DCAN effectively isolates diagnostically relevant regions while mitigating the impact of noise and lighting variations, common challenges in polyp detection. Our results demonstrate that DCAN outperforms traditional attention methods, offering improved accuracy, and explainability. The integration of adaptive attention refinement ensures robust localization without explicit localization modules, paving the way for more reliable and interpretable colorectal cancer diagnostics. Future research will explore multi-modal learning frameworks that integrate endoscopic images with text data, enabling medical report generation. 

\printbibliography[title={References}]

\end{document}